\documentclass{bmvc2k}

\usepackage{tabu}
\usepackage{multirow}

\usepackage{wrapfig}
\usepackage{algorithm}
\usepackage{algorithmic}
\usepackage[final]{pdfpages}

\usepackage{amsmath,amsfonts,bm}
\usepackage{mathtools, cuted}

\def\rvt{{\mathbf{t}}}

\def\rvx{{\mathbf{x}}}

\def\rvz{{\mathbf{z}}}

\def\vr{{\bm{r}}}
\def\vs{{\bm{s}}}
\def\vt{{\bm{t}}}

\def\vv{{\bm{v}}}
\def\vw{{\bm{w}}}
\def\vx{{\bm{x}}}

\def\vz{{\bm{z}}}

\DeclareMathOperator*{\argmax}{arg\,max}

\newcommand{\E}{\mathbb{E}}
\usepackage{stackengine}
\def\delequal{\mathrel{\ensurestackMath{\stackon[1pt]{=}{\scriptscriptstyle\Delta}}}} 
\newcommand{\KL}{D_{\mathrm{KL}}}

\title{LaDDer: Latent Data Distribution Modelling with a Generative Prior}

\addauthor{Shuyu Lin}{https://shuyulin.co.uk/}{1}
\addauthor{Ronald Clark}{http://www.ronnieclark.co.uk/}{2}

\addinstitution{
 Computer Science Department,\\
 University of Oxford,\\
 Oxford, UK
}
\addinstitution{
 Imperial College London\\
 South Kensington, \\
 London, UK
}

\runninghead{Lin and Clark}{LaDDer - Latent Data Distribution Modelling}

\usepackage{hyperref}
\hypersetup{
    colorlinks=true,
    linkcolor=blue,
    filecolor=magenta,      
    urlcolor=cyan,
}

\begin{document}

\maketitle

\begin{abstract}
In this paper, we show that the performance of a learnt generative model is closely related to the model's ability to accurately represent the inferred \textbf{latent data distribution}, i.e. its topology and structural properties. 
We propose LaDDer to achieve accurate modelling of the latent data distribution in a variational autoencoder framework and to facilitate better representation learning. The central idea of LaDDer is a meta-embedding concept, which uses multiple VAE models to learn an embedding of the embeddings, forming a ladder of encodings. We use a non-parametric mixture as the hyper prior for the innermost VAE and learn all the parameters in a unified variational framework. From extensive experiments, we show that our LaDDer model is able to accurately estimate complex latent distribution and results in improvement in the representation quality. We also propose a novel latent space interpolation method that utilises the derived data distribution. The code and demos are available at \url{https://github.com/lin-shuyu/ladder-latent-data-distribution-modelling}. 


\end{abstract}

\section{Introduction}

\label{sec:intro}

Variational autoencoders (VAEs) \cite{VAE-1,VAE-2} are probabilistic latent variable models that aim to learn rich representations from large amounts of data in an unsupervised manner. A trained VAE consists of a generative decoder that generates a data sample from a latent code and a variational encoder that maps a data sample to an approximate posterior distribution over latent variables. Amongst different types of generative models including GANs \cite{GAN}, flow models \cite{NICE,realNVP} and autoregressive models \cite{Nade}, VAEs have been favoured for their training stability, strong theoretical grounding and ability to learn well-structured latent representations. In general, the quality of a learned generative model is dependent on two factors: 1) the quality of the inferred data distribution, which defines the topology and structural properties of the \textbf{latent space}; 2) the ability to generate good quality samples in the \textbf{data space}.

Thanks to all the preferable properties, VAEs have been widely applied in many computer vision applications, including image synthesis \cite{image-synthesis-eg1,image-synthesis-eg2,image-synthesis-eg3}, human motion modelling \cite{human-motion-modelling1,human-motion-modelling2} and 3D reconstruction \cite{3D-recons-eg2,codeslam,czarnowski2020deepfactors}. However, VAEs still struggle when modelling complex data such as images. 
Specifically, compared to other models, the generated images of VAEs can be somewhat blurry, lack of fine details and have limited diversity. An entire research field is dedicated to improving VAE learning quality. In this paper, we focus on a key component of VAE learning objective that involves restricting the learned latent space representation to a chosen prior probability. As we cannot observe the true latent distribution from which the data were generated, it is impossible to choose a perfect prior distribution a priori. Therefore, many works, which adopt a simple and inflexible prior in the form of a unit Gaussian distribution, often suffer from an over-regularised latent representation, as the model tries to get the encoder to shoehorn the data to fit the simple prior, sacrificing the quality of generated images. This suggests that using a more flexible prior distribution, or treating the prior as a parameterised model, can reduce or eliminate the over-regularisation issue and lead to improvement in the representation quality. 

\begin{figure}[t]
    \centering
    \includegraphics[width=0.9\textwidth]{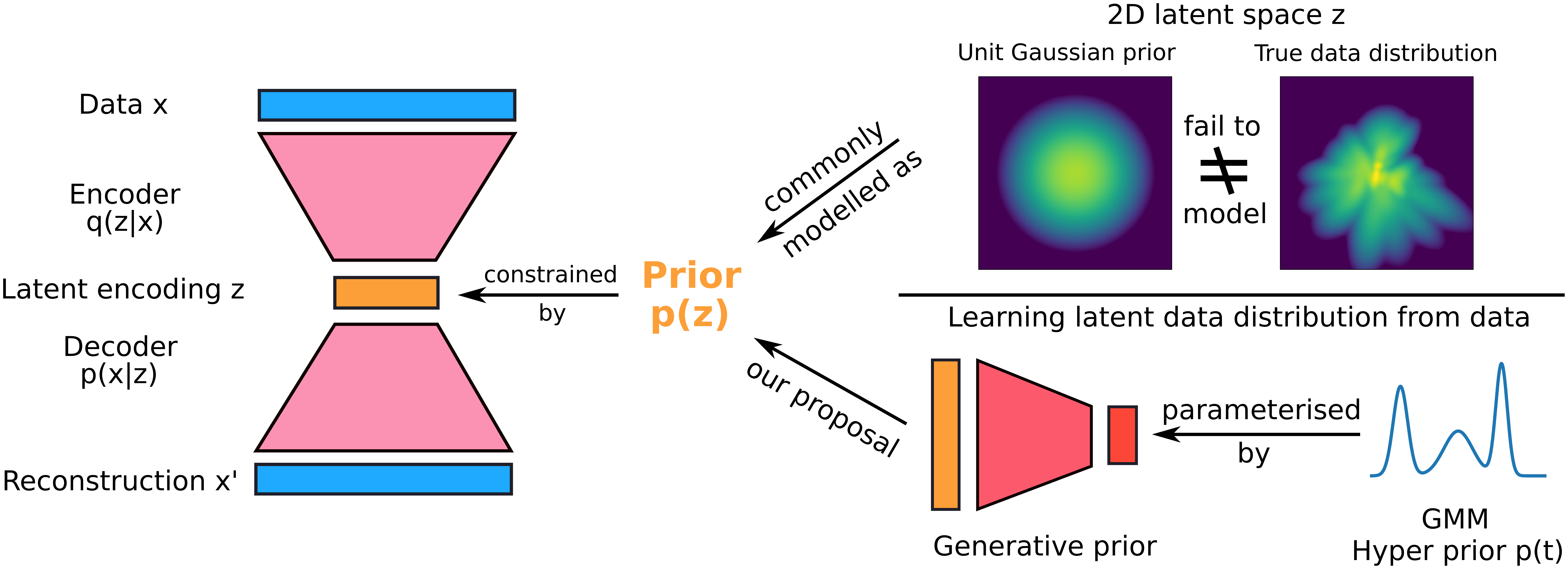}
    \vspace{-2mm}
    \caption{\textbf{LaDDer's key concept}. LaDDer adopts a generative prior, which consists of a mixture hyper prior and a series of VAEs each acting on its predecessor's latent encodings.}
    \label{fig:system_overview}
    \vspace{-4mm}
\end{figure}

Following these insights, we propose LaDDer, a method that allows us to accurately model the prior distribution in a VAE framework. The central idea of our method is a meta-embedding concept which, in short, derives a latent embedding of a latent embedding.
Specifically as shown in Fig.~\ref{fig:system_overview}, our approach consists of multiple VAE models each acting on its predecessor's latent representation and forming a ladder of encodings. We use a non-parametric mixture as the hyper prior for the innermost VAE. The hyper prior together with the intermediate encoders forms a \textbf{generative prior} for the outermost VAE. We learn the parameters of all the VAE networks, along with the non-parametric mixture, in a unified variational inference framework. From extensive experiments, we show that our LaDDer model is able to accurately estimate complex \textbf{latent data distribution} and results in improvement in the representation quality. We also propose a \textbf{novel latent space interpolation} method and show how to best utilise the derived latent distribution in further tasks.  

\section{Background}
\label{sec:background}
Here we introduce VAE models and explain why the prior distribution is crucial in producing good learning outcomes. Given a dataset of N observations $\mathcal{D}_N = \{\vx_1, \cdots, \vx_N\}$, VAEs assume all data samples $\vx_i$ are generated from a low-dimensional latent space $\rvz$ under a latent variable model $p_{\theta}(\rvx, \rvz) = p_{\theta}(\rvx | \rvz)p(\rvz)$, where $p_{\theta}(\rvx | \rvz)$ denotes the generative model (decoder) parameterised by $\theta$. VAEs learn the model parameters $\theta$ by maximising the marginal log likelihood for all data points in $\mathcal{D}_N$, i.e. 
$\argmax_{\theta} \E_{p_{\mathcal{D}}(\rvx)} [\log p_{\theta}(\rvx)]$,  
where $p_{\mathcal{D}}(\rvx)$ is the empirical data distribution and $p_{\theta}(\rvx)=\int p_{\theta}(\rvx|\rvz)p(\rvz) \textrm{d}\rvz$. However, directly evaluating the marginal log likelihood is often not feasible, as integration over the network $p_{\theta}(\rvx|\rvz)$ is not trivial.
To obtain an analytical learning objective, VAEs \cite{VAE-1,VAE-2} use variational inference and derive an evidence lower bound (ELBO) $\mathcal{L}(\rvx; \theta, \phi)$ to the marginal log likelihood, i.e. 
$
    \E_{p_{\mathcal{D}}(\rvx)}
    [\log p_{\theta}(\rvx)]  
    \,\geq \, \mathcal{L}(\rvx; \theta, \phi) \label{eq:VAE-ELBO-obj1}
$, as shown in Eq (\ref{eq:VAE-ELBO-obj2}):
\begin{align}
    \mathcal{L}(\rvx; \theta, \phi) & \delequal \,
    \E_{p_{\mathcal{D}}(\rvx)}\big[
    \E_{q_{\phi}(\rvz|\rvx)}[\log p_{\theta}(\rvx|\rvz)]
    \;-\;
    \KL[\,q_{\phi}(\rvz|\rvx) \Vert p(\rvz)]
    \big] \label{eq:VAE-ELBO-obj2} \\
    \delequal  \, \E_{p_{\mathcal{D}}(\rvx)}&\big[ 
    \underbrace{\E_{q_{\phi}(\rvz|\rvx)}[\log p_{\theta}(\rvx|\rvz)]}
    _\text{\small \textcircled{1} \textrm{\scriptsize{reconstruction likelihood}}}
    \;-\;
    \underbrace{\E_{q_{\phi}(\rvz|\rvx)}[\log q_{\phi}(\rvz|\rvx)]}
    _\text{\small \textcircled{2} \textrm{\scriptsize{posterior entropy}}}
    \;+
    \underbrace{\E_{q_{\phi}(\rvz|\rvx)}[\log p(\rvz)]}
    _\text{\small \textcircled{3} \textrm{\scriptsize{cross-entropy wrt prior}}} \big].
    \label{eq:VAE-ELBO-obj3} 
\end{align}

By breaking down the KL divergence term in Eq (\ref{eq:VAE-ELBO-obj2}) using the definition of KL, we obtain an ELBO expression in Eq (\ref{eq:VAE-ELBO-obj3}). 
Here we see that the ELBO contains three terms: \textcircled{\small{1}} a reconstruction likelihood that encourages good reconstruction through the auto-encoding process, \textcircled{\small{2}} a negative posterior entropy that favours $q_{\phi}(\rvz|\rvx)$ with large variances and finally \textcircled{\small{3}} a cross-entropy between the posterior $q_{\phi}(\rvz|\rvx)$ and the prior $p(\rvz)$ that regularises the posteriors to comply with the target prior distribution. We can re-arrange \textcircled{\small{3}} into a cross-entropy between the aggregate posterior $q_{\phi}(\rvz)$ and the prior $p(\rvz)$, as shown below:
\begin{gather} \label{eq:rearrange-cross-entropy}
    \E_{p_{\mathcal{D}}(\rvx)}\E_{q_{\phi}(\rvz|\rvx)}[\log p(\rvz)]
    = \E_{q_{\phi}(\rvz)} [\log p(\rvz)], \;\;
    \textrm{where} \;\; 
    q_{\phi}(\rvz) 
    =
    \E_{p_{\mathcal{D}}(\rvx)} [q_{\phi}(\rvz|\rvx)]. 
\end{gather}
This re-arrangement reveals that term \textcircled{\small{3}} in the ELBO loss encourages the prior $p(\rvz)$ and the inferred latent data distribution $q_{\phi}(\rvz)$ to match each other. If an overly limiting prior is used, then the learnt data distribution will diverge from the true data distribution. Details of the re-arrangement are given in Supplementary Materials (SM) A1. 

\vspace{-2mm}
\subsection{Related work}

Many approaches have been proposed to improve VAE's modelling ability for rich, high-dimensional data. For example, \cite{expressive-post-1,expressive-post-2,expressive-post-3,expressive-post-4} propose to use more flexible sample posteriors $q_{\phi}(\rvz|\vx)$ to increase the encoder's expressive power and improve the model's performance in all three loss terms of Eq (\ref{eq:VAE-ELBO-obj3}). \cite{IWAE,factorVAE,beta-VAE} focus on deriving alternative learning objectives in order to produce representations with preferred qualities, such as disentanglement. 

We take a different approach from these lines of research. In this work, we focus on modelling the prior distribution $p(\rvz)$ accurately, which then results in improved learning performance and facilitates further tasks, such as latent space interpolation. Several works have considered using more flexible distributions than a unit Gaussian as priors, such as a stick-breaking prior \cite{SBP-prior}, a Chinese Restaurant Process prior \cite{CRP-prior} or a Gaussian mixture (GM) prior \cite{deep-clustering,VampPrior,SVAE,alpha-VAE}. 
Those methods often have limited performance for complex datasets where high-dimensional latent space is needed to facilitate the generative modelling.
A closely related work is \cite{hierarchical-prior}, where a generative model is used to parameterise the prior. Our method differs from \cite{hierarchical-prior}, as we realise that the introduction of a generative prior alone does not guarantee accurate modelling of the latent data distribution. Thus, we introduce a GM hyper prior to complete the modelling task. In addition, we optimise all the model parameters in a coherent lower bound objective, whereas \cite{hierarchical-prior} has to set up a constrained optimisation to replace the ELBO objective in order to learn the model parameters.    

Another line of research in VAEs is devoted to improving the generation quality, leading to the impressive image quality shown in VQ-VAE2 \cite{VQ-VAE2}. A major difference between VQ-VAE2 and our work lies in the different approaches taken to model the prior distribution. VQ-VAE2 learns an auto-regressive prior (using a pixelCNN model \cite{PixelCNN}) in a post-hoc, 2nd stage (after the autoencoder is trained). Our model trains both autoencoder and prior modules jointly under the same objective. Further, VQ-VAE2 employs a 2D latent representation, where a single feature vector corresponds to a local patch of the generated image. It is extremely difficult to manipulate such a 2D representation to generate globally consistent changes. In contrast, we adopt a global code to represent the whole image. We can easily traverse across the derived data manifold and generate smoothly changing data samples.

\section{Our Method}
\label{sec:proposal}
As shown in Fig.~\ref{fig:system_overview}, we propose a generative prior, which includes an additional VAE model to project the original data VAE's encodings to an even lower dimensional space and a hyper prior for this prior VAE. We parameterise the hyper prior as a Gaussian mixture model (GMM). This design allows us to accurately estimate the latent data distribution, as the generative prior is sufficiently flexible to fit any arbitrarily complex distribution. At the same time, the optimisation of our LaDDer model can be easily integrated into the VAE ELBO objective, which we will demonstrate in this section. In the end, we also demonstrate how to use the latent data distribution derived from our LaDDer learning to facilitate a latent space interpolation task.

\vspace{-2mm}
\subsection{The VAE Unit in Our Generative Prior}
\label{sec:generative-prior}
We first introduce the VAE unit in our generative prior.
Similar to the VAE for data samples introduced in Section \ref{sec:background}, the prior VAE is also formulated by a latent variable model 
$p_{\alpha}(\rvz, \rvt) = p_{\alpha}(\rvz|\rvt) p(\rvt)$,
which governs the generation of latent encodings $\vz_i$ through 1) a hyper prior $p(\rvt)$ that resides in an even lower dimensional space and 2) an encoding decoder $p_{\alpha}(\rvz|\rvt)$ which is parameterised by a neural network with parameters $\alpha$.
To optimise the prior VAE, we can introduce a variational encoder $q_{\beta}(\rvt|\rvz)$ parameterised by $\beta$ and learn both $\alpha$ and $\beta$ by maximising an ELBO objective $\mathcal{L}(\rvz; \alpha, \beta)$ similar to Eq (\ref{eq:VAE-ELBO-obj3}) for this prior model, i.e.
$ \label{eq:inner-VAE-ELBO}
    \E_{q_{\phi}(\rvz|\rvx)} [\log p_{\alpha}(\rvz)]  
    \geq 
    \mathcal{L}(\rvz; \alpha, \beta) 
$, where:
\begin{align}
    \mathcal{L}(\rvz; \alpha, \beta) &\delequal
    \E_{q_{\phi}(\rvz|\rvx)}\big[
    \E_{q_{\beta}(\rvt|\rvz)}[\log p_{\alpha}(\rvz|\rvt)]
    -\E_{q_{\beta}(\rvt|\rvz)}[\log q_{\beta}(\rvt|\rvz)]
    +\E_{q_{\beta}(\rvt|\rvz)}[\log p(\rvt)] 
    \big]. 
    \label{eq:inner-VAE-ELBO2}
\end{align}

Notice that $\mathcal{L}(\rvz; \alpha, \beta)$ is a lower bound to the likelihood $\E_{q_{\phi}(\rvz|\rvx)} [\log p_{\alpha}(\rvz)]$, which is equivalent to the cross-entropy term \textcircled{\small{3}} in Eq (\ref{eq:VAE-ELBO-obj3}). This connection allows us to integrate the learning objective of this prior VAE, i.e. $\mathcal{L}(\rvz; \alpha, \beta)$, into the ELBO for the original data VAE, i.e. $\mathcal{L}(\rvx; \theta, \phi)$, and obtain a new lower bound $\mathcal{L}'(\rvx; \theta, \phi, \alpha, \beta)$ to the data ELBO $\mathcal{L}(\rvx; \theta, \phi)$, i.e.
$ 
    \mathcal{L}(\rvx; \theta, \phi)
    \geq \mathcal{L}'(\rvx; \theta, \phi, \alpha, \beta) 
$, where:
\begin{align}
    \mathcal{L}'(\rvx; \theta, \phi, \alpha, \beta)=
    \E_{p_{\mathcal{D}}(\rvx)} \big[
    \E_{q_{\phi}(\rvz|\rvx)}[\log p_{\theta}(\rvx|\rvz)]
    -\E_{q_{\phi}(\rvz|\rvx)}[\log q_{\phi}(\rvz|\rvx)]
    +\mathcal{L}(\rvz; \alpha, \beta)
    \big].
    \label{eq:prior-lower-bound-to-ELBO}
\end{align}

\vspace{-4mm}
\subsection{Variational Gaussian Mixture Model for the Hyper Prior}
\label{sec:hyper-prior}
The prior VAE unit defined in Section \ref{sec:generative-prior} introduces a hyper prior $p(\rvt)$. From Eq (\ref{eq:rearrange-cross-entropy}), we know that the optimal hyper prior should be matched to the aggregate hyper posterior, i.e. $p(\rvt) \approx q_{\beta, \phi}(\rvt)$. 
To facilitate the matching, we parameterise $p(\rvt)$ with a Gaussian mixture model (GMM) of M components (M $\ll$ N):
$
    p(\rvt)
    = 
    \sum_{m=1}^{M}w_m \mathcal{N}(\rvt; \,\mu_m, \,\Sigma_m),
$
where $w_m$ is the weight for each Gaussian mixture ($w_m > 0$ and $\sum_m w_m =1$) and $\mu_m$ and $\Sigma_m$ are the mean and covariance matrix for the $m$-th Gaussian mixture. To fit $p(\rvt)$ to $q_{\beta, \phi}(\rvt)$, we resort to variational inference techniques introduced in \cite{VGMM}. 
Here we give a high-level sketch of the algorithm and we refer interested readers to \cite{VGMM} for further mathematical details. Firstly, we define the generative model of the GMM by introducing the following distributions to the GMM parameters $(\vw, \mu, \Sigma)$ and all N prior encoding samples $\vt_i$ (the encodings of $\vz_i$):
$ \label{eq:generative-distribution1}
    w_m \sim \mathcal{B}eta(1, \alpha_0), \;\; 
    \mu_m \sim \mathcal{N}(0, \mathcal{I}), \;\;
    \Sigma_m \sim \mathcal{W}(d_t, \mathcal{I}),\;\;
    k_i \sim \textrm{Cat}(\vw),\;\;
    \vt_i \sim \mathcal{N}(\mu_{k_i}, \Sigma_{k_i}),
$
where $k_i$ indicates the choice of mixture components for $i$-th sample $\vt_i$. 
Secondly, we introduce the following variational distributions under the mean-field assumption for all the model variables $W=(\vw, \mu, \Sigma, k_i)$:
$ \label{eq:variational-distribution}
    w_m \sim \mathcal{B}eta(\gamma_{m,1}, \gamma_{m,2}), \;\; 
    \mu_m \sim \mathcal{N}(\vv_m, \mathcal{I}), \;\;
    \Sigma_m \sim \mathcal{W}(a_m, \mathcal{B}_m), \;\;
    k_i \sim \textrm{Discrete}(\vr_i),
$
where $\xi = (\gamma_{m,1}, \gamma_{m,2}, \vv_m, a_m, \mathcal{B}_m, \vr_i)$ for $m=1,\cdots,M$ and $i=1,\cdots,N$ denotes variational variables and needs to be optimised. Thirdly, a variational bound on the log likelihood of $p(\rvt)$ is introduced as the learning objective to optimise $\xi$, as shown below:
\begin{align} \label{eq:VGMM-bound}
    \log p(\rvt | \alpha_0, M) 
    \geq
    \mathcal{L}(\rvt; \xi)
    =
    \E_{q_{\xi}(W)} [\log p(W, \rvt| \alpha_0, M)]
    -\E_{q_{\xi}(W)} [\log q_{\xi}(W)],
\end{align}
where $p(W, \rvt| \alpha_0, M)$ denotes the product of all the generative distributions introduced in step 1 and $q_{\xi}(W)$ denotes the product of all the variational distributions introduced in step 2.

Now substituting $\mathcal{L}(\rvt; \xi)$ in Eq (\ref{eq:VGMM-bound}) into Eq (\ref{eq:inner-VAE-ELBO2}) to replace $\log p(\rvt)$, we obtain a lower bound $\mathcal{L}'(\rvz; \alpha, \beta, \xi)$ to the prior ELBO $\mathcal{L}(\rvz; \alpha, \beta)$, i.e. $\mathcal{L}(\rvz; \alpha, \beta) \geq \mathcal{L}'(\rvz; \alpha, \beta, \xi)$, as
\begin{align} \label{eq:LB-prior-ELBO}
    \mathcal{L}'(\rvz; \alpha, \beta, \xi) 
    =
    \E_{q_{\phi}(\rvz|\rvx)}\big[
    \E_{q_{\beta}(\rvt|\rvz)}[\log p_{\alpha}(\rvz|\rvt)]
    -\E_{q_{\beta}(\rvt|\rvz)}[\log q_{\beta}(\rvt|\rvz)]
    + \E_{q_{\beta}(\rvt|\rvz)} [\mathcal{L}(\rvt; \xi)]
    \big]. 
\end{align}
Now substituting $\mathcal{L}'(\rvz; \alpha, \beta, \xi)$ into the lower bound $\mathcal{L}'(\rvx; \theta, \phi, \alpha, \beta)$ in Eq (\ref{eq:prior-lower-bound-to-ELBO}) to replace $\mathcal{L}(\rvz; \alpha, \beta)$, we obtain a final lower bound $\mathcal{L}''(\rvx; \Theta)$ to the data likelihood $\E_{p_{\mathcal{D}}(\rvx)} [\log p_{\theta}(\rvx)]$, i.e. 
$ 
    \E_{p_{\mathcal{D}}(\rvx)} [\log p_{\theta}(\rvx)]  
    \geq 
    \mathcal{L}''(\rvx; \Theta)
$, where $\Theta = [\theta, \phi, \alpha, \beta, \xi]$ denotes all the parameters in the data VAE, prior VAE and GMM hyper prior, and
\begin{align}
\label{final-LB}
    \mathcal{L}''(\rvx; \Theta)=
    \E_{p_{\mathcal{D}}(\rvx)} \big[
    \E_{q_{\phi}(\rvz|\rvx)}[\log p_{\theta}(\rvx|\rvz)]
    -\E_{q_{\phi}(\rvz|\rvx)}[\log q_{\phi}(\rvz|\rvx)]
    +\mathcal{L}'(\rvz; \alpha, \beta, \xi)
    \big].
\end{align}






\vspace{-4mm}
\subsection{Block Coordinate Ascent to Optimise Model Parameters}
\label{sec:BCA-optimisation}
With the new lower bound in Eq (\ref{final-LB}), we are ready to introduce our block coordinate ascent algorithm which optimises all the model parameters $\Theta = [\theta, \phi, \alpha, \beta, \xi]$. Notice that the GMM hyper prior's parameter $\xi$ are only involved in the term $\mathcal{L}(\rvt; \xi)$ from Eq (\ref{eq:VGMM-bound}). We can update $\xi$ by maximising $\mathcal{L}(\rvt; \xi)$ alone. With the newly updated $\xi$, we can then update the prior VAE's parameter $\alpha, \beta$, which only present in the prior VAE's ELBO $\mathcal{L}'(\rvz; \alpha, \beta | \xi)$ from Eq~(\ref{eq:LB-prior-ELBO}). Finally, with $\xi, \alpha, \beta$ updated, we can now update the data VAE's parameters $\theta, \phi$ by maximising $\mathcal{L}''(\rvx; \theta, \phi | \alpha, \beta, \xi)$ from Eq~(\ref{final-LB}) with $\alpha, \beta, \xi$ being fixed. An algorithmic illustration for this optimisation procedure is given in SM B. Notice that we initialise the GMM hyper prior as an uninformative unit Gaussian to begin the optimisation. 

\vspace{-2mm}
\subsection{Shortest Likely Path to Traverse the Data Manifold}
\label{sec:SLP}
In Section \ref{sec:generative-prior}-\ref{sec:BCA-optimisation}, we introduce a generative prior and an optimisation scheme to facilitate accurate modelling of the latent data distribution. Now we demonstrate how to use the derived latent distribution in a latent space traversal task, where a path along the learnt data manifold needs to be inferred to interpolate between two data samples. 
Such interpolation has been used commonly in previous works \cite{factorVAE,Glow,UnsupervisedRL,AAR} to illustrate the smoothness of the learnt data manifold. Traditionally, the traversal is done by a linear interpolation between the encodings of a pair of query images, which we refer as shortest path (SP) interpolation. This method ignores the nonlinear topology of the data manifold in the latent space and often results in poor interpolated images, which are generated faraway from the data manifold.  

To achieve better interpolation results, we propose to formulate the traversal task as an optimisation problem which aims to find an optimal path that is shortest while remaining close to the data manifold. We refer to our interpolation scheme as shortest likelihood path (SLP) interpolation and design an optimisation objective as follows:
\begin{align} \label{eq:SLP-obj}
    \mathcal{O}_{\textrm{SLP}} 
    \;\;=\;\;
    \underbrace{\textrm{L}_{\textrm{path}}(\vs)}
    _\text{\textrm{\scriptsize{path length}}}
    \;\;+\;\;
    \underbrace{\textrm{std}_{\textrm{step length}}(\vs)}
    _\text{\textrm{\scriptsize{equal step length}}}
    \;\;-\;\;
    \underbrace{\log p(\rvt=\vs),}
    _\text{\textrm{\scriptsize{path likelihood}}}
\end{align}
where $\vs=[s_1, \cdots, s_J]^T$ denotes a list of J steps along the traversal path and each $s_j$ denotes a single encoding that can be projected back to the data space as an interpolated image. As can be seen, our $\mathcal{O}_{\textrm{SLP}}$ objective contains three terms: 1) $\textrm{L}_{\textrm{path}}(\vs)$ minimises the current path length, 2) $\textrm{std}_{\textrm{step length}}(\vs)$ requires steps to be evenly distributed along the path and 3) $\log p(\rvt=\vs)$ ensures all the interpolated encodings to be generated from the inferred data distribution $p(\rvt)$. To find an optimal path $\vs$, we minimise $\mathcal{O}_{\textrm{SLP}}$ wrt $\vs$ by a standard optimisation scheme, such as AdamOptimiser \cite{AdamOpt} that is used in our experiments. 
Notice that our SLP interpolation can be generally applied to any representation learning algorithm, as long as the likelihood of the latent data distribution $p(\rvz)$ can be easily evaluated. 

\section{Experiments and Results}
\label{sec:result}
We carry out extensive experiments on MNIST \cite{MNIST}, Fashion MNIST \cite{fashion-mnist} and CelebA \cite{CelebA-Dataset} datasets to evaluate our LaDDer model. For all datasets, images are treated as real-valued data and we use a Laplace distribution to model the decoder $p_{\theta}(\rvx|\rvz)$. The corresponding reconstruction likelihood (term \textcircled{\small{1}} in Eq~(\ref{eq:VAE-ELBO-obj3})) is derived in SM C, following \cite{alpha-VAE}. We compare our method to 4 other approaches, including the original VAE with a normal prior \cite{VAE-1,VAE-2}, VAE with a GMM prior \cite{deep-clustering, SVAE,alpha-VAE}, VAE with a hierarchical prior \cite{hierarchical-prior} where a generative prior with a normal hyper prior is used, and VampPrior \cite{VampPrior} where the prior is modelled as an average encoding of a set of inferred pseudo-inputs. More results, details of data pre-processing and model architectures are given in SM D-F.

\vspace{-2mm}
\subsection{Our Method Better Estimates the Latent Data Distribution}
\begin{figure*}[t]
    \centering
    \includegraphics[width=0.95\textwidth]{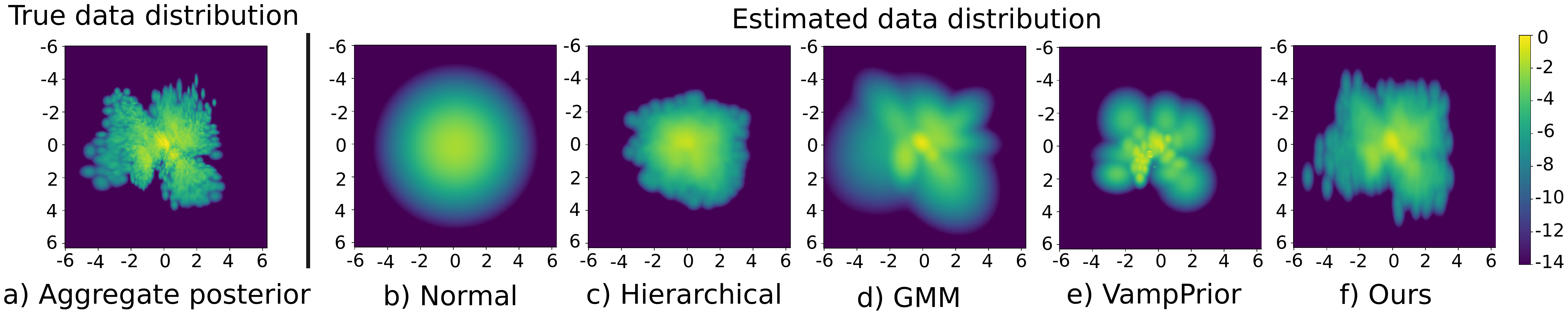}
    \vspace{-2mm}
    \caption{\textbf{Visualising the modelling of MNIST data distribution in a 2D latent space}. We first train a VAE model with the unit Gaussian prior. Then we fix the autoencoder model and fit different prior models to the derived encodings. The true data distribution (aggregate posterior $q_{\phi}(\rvz)$) is shown in (a), where we encode 10k training images and visualise their posteriors. In b-f, we visualise the pdfs of the 5 different prior methods. Our generative prior produces the best fit to the true data distribution.}
    \label{fig:mnist-2d-latent-space}
    \vspace{-1mm}
\end{figure*}

We first show that our LaDDer model can achieve a better modelling of the latent data distribution, i.e. more accurate estimation of the aggregate posterior $q_{\phi}(\rvz)$.
Fig.~\ref{fig:mnist-2d-latent-space} visualises the aggregate posterior $q_{\phi}(\rvz)$ and the optimised priors from 5 different approaches, where VAE models are trained for 2D latent space on MNIST dataset. 
The true data distribution shown in Fig.~\ref{fig:mnist-2d-latent-space}a contains complex low density regions, which correspond to natural boundaries between different object classes. 
An inflexible prior model, such as a unit Gaussian (Fig.~\ref{fig:mnist-2d-latent-space}b), over-represents such low density regions.
The hierarchical prior (Fig.~\ref{fig:mnist-2d-latent-space}c) also fails to model $q_{\phi}(\rvz)$ accurately, because the prior VAE is overly regularised by its inflexible hyper prior. The other three methods (Fig.~\ref{fig:mnist-2d-latent-space}d-f), where the prior $p(\rvz)$ is parameterised with mixture models, obtain better fitting. The difference across these three methods lies in the number of mixture components. 
Both the GMM prior and the VampPrior have a finite number of mixture components, whereas our LaDDer prior contains an infinite number of mixture components, as our prior is estimated by integrating over all $t$-values: $p_{\alpha}(\rvz)=\int p_{\alpha}(\rvz|\rvt)p(\rvt)\,\textrm{d}\rvt$. As a result, our method is able to smoothly fit any arbitrarily complex $q_{\phi}(\rvz)$ with no artificial boundaries as the ones introduced in VampPrior.

\begin{figure*}[t]
    \centering
    \includegraphics[width=\textwidth]{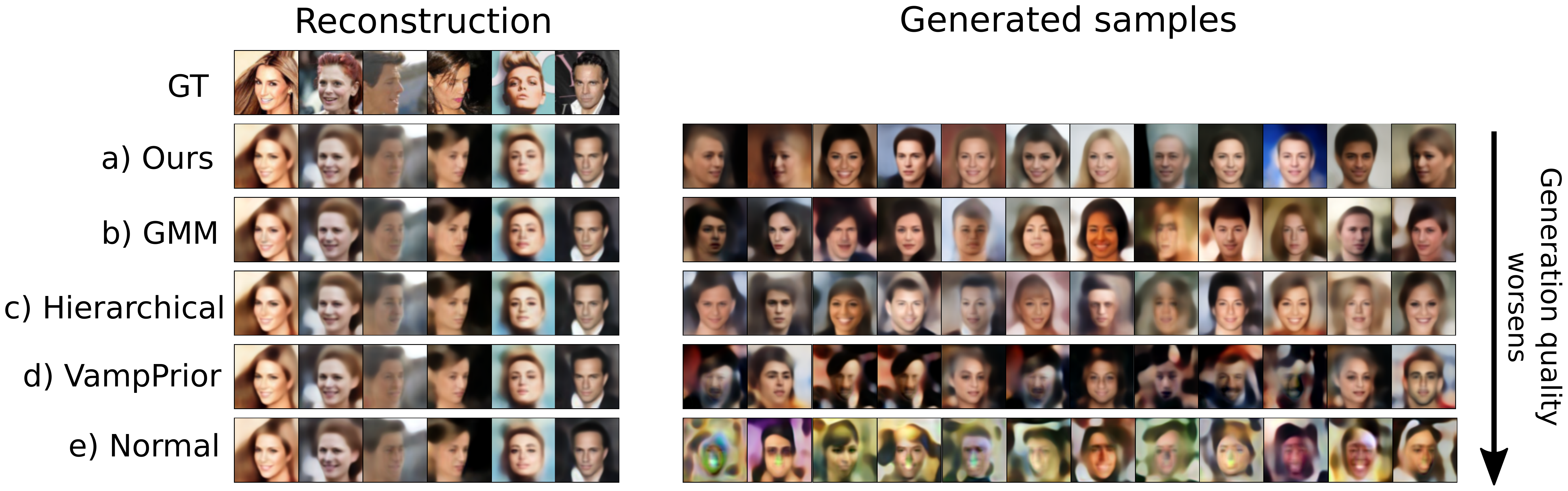}
    \vspace{-2mm}
    \caption{\textbf{Comparison of reconstruction and generation quality across different VAE models}. All five models trained with celebA dataset can produce similar quality reconstruction, but the generation quality varies significantly samples. The difference between reconstruction and generation quality indicates the importance of a prior model in modelling the data distribution and generating good quality samples. 
    Our generative prior can model the data distribution $q_{\phi}(\rvz)$ well, hence generating best quality samples with minimal gap to the quality of reconstructed samples. More examples are in Supplementary Materials F1.}
    \label{fig:generation-quality-comparison}
    \vspace{-3mm}
\end{figure*}

Having a prior that can better fit $q_{\phi}(\rvz)$ also leads to improved generation quality. Fig.~\ref{fig:generation-quality-comparison} gives examples of both reconstructed and generated samples from the 5 VAE models trained under the CelebA dataset. All 5 methods can produce good reconstruction, but the generation quality varies significantly. This indicates that the bad generation quality commonly reported in VAE models does not merely result from the autoencoder architecture, but is also a result of the prior $p(\rvz)$ failing to represent the topology of the learned latent data distribution. Generating from the prior that does not represent the true data distribution unsurprisingly produce unrealistic samples. Our proposed generative prior addresses this issue and produces generation quality that is almost as good as the model's reconstruction quality. The improved generation quality is supported by the FID score \cite{FID1}, which is a quantitative measure to evaluate visual quality of image samples, shown in Table \ref{tab:celeba-FID}.   
\newcolumntype{A}{ >{\centering\arraybackslash} m{1.2cm} }
\newcolumntype{B}{ >{\centering\arraybackslash} m{1.5cm} }
\begin{table}[t]
    \centering
    \scriptsize
    \caption{\textbf{FID scores} (lower is better) of generated and reconstructed samples. Our LaDDer model achieves the best sample quality and the minimal gap between the two.}
    \vspace{1mm}
    \begin{tabular}{|B|A|A|A|A|A|}
        \hline
                         & Normal           & VampPrior           & GMM             & Hierarchical             & Ours       \\ \hline
        Generation         & 250.3 $\pm$ 1.5  & 182.9 $\pm$ 1.2     & 140.4 $\pm$ 0.9 & 143.8 $\pm$ 1.0        & \textbf{132.7} $\pm$ 0.8     \\ \hline
        Reconstruction     & 100.3 $\pm$ 0.6  & 95.5  $\pm$ 0.7     & 102.3 $\pm$ 0.5 & 99.7  $\pm$ 0.6        & \textbf{95.3}  $\pm$ 0.5     \\ \hline
        Difference         & 150.0 $\pm$ 2.1  & 87.4  $\pm$ 1.9     & 38.1  $\pm$ 1.4 & 44.1  $\pm$
        1.6        & \textbf{37.4}  $\pm$ 1.3     \\ \hline
    \end{tabular}
    \label{tab:celeba-FID}
\end{table}

We also evaluate the cross-entropy $E_{q_{\phi}(\rvz)}[\log p(\rvz)]$ (\textcircled{\small{3}} in Eq (\ref{eq:VAE-ELBO-obj3})) between the aggregate posterior $q_{\phi}(\rvz)$ and the prior $p(\rvz)$ to measure the level of matching between the two distributions.
Higher $E_{q_{\phi}(\rvz)}[\log p(\rvz)]$ indicates a better fit of the prior $p(\rvz)$ to the learnt data distribution $q_{\phi}(\rvz)$. 
As shown in Table \ref{tab:cross-entropy}, GMM priors can achieve very good fitting for relatively simple datasets, such as MNIST. However, when the dataset becomes more complex and the required latent dimension increases, such as fashion-MNIST and CelebA datasets, our generative prior, which is more flexible to fit complex distributions, obtains the highest cross-entropy and achieve the best fit. 
\newcolumntype{C}{ >{\centering\arraybackslash} m{1.4cm} }
\newcolumntype{D}{ >{\centering\arraybackslash} m{1.6cm} }
\begin{table}[t]
    \centering
    \scriptsize
    \caption{\textbf{Cross-entropy} $E_{q_{\phi}(\rvz)}[\log p(\rvz)]$ (higher is bettter) between the aggregate posterior and the prior. Our LaDDer model achieves the best fit to $q_{\phi}(\rvz)$ in more complex datasets.}
    \vspace{1mm}
    \begin{tabular}{|D|C|C|C|C|C|}
        \hline
        Dataset         & Normal            & VampPrior         & GMM                        & Hierarchical & Ours                            \\ \hline
        MNIST           & -19.0 $\pm$ 0.1   & -16.2 $\pm$ 0.1   & \textbf{-8.4} $\pm$ 0.9    & -17.5 $\pm$ 0.1    & -12.3 $\pm$ 0.1           \\ \hline
        fashion-MNIST   & -52.5 $\pm$ 0.6   & -60.8 $\pm$ 1.0   & -45.2 $\pm$ 0.8            & -63.8 $\pm$ 0.9    &  \textbf{-32.8} $\pm$ 0.7 \\ \hline
        CelebA          & -351.4 $\pm$ 4.3  & -380.7 $\pm$ 3.4  & -290.6 $\pm$ 54.8          & -358.0 $\pm$ 7.3    & \textbf{-71.9} $\pm$ 2.1 \\ \hline
    \end{tabular}
    \vspace{-3mm}
    \label{tab:cross-entropy}    
\end{table}

\vspace{-2mm}
\subsection{Overall Generative Modelling Performance Improves}
A more flexible prior $p(\rvz)$ also leads to better generative modelling performance, in terms of better reconstruction quality and higher ELBO objective. This effect is clearly shown in Table \ref{tab:ELBO-recons-error}, where our method produces the highest ELBO value and the lowest reconstruction error across all datasets. All models are trained under the same conditions and the evaluation is repeated for 5 times to estimate the variance of the results.   

\newcolumntype{E}{ >{\centering\arraybackslash} m{0.9cm} }
\begin{table}[h]
    \centering
    \scriptsize
    \caption{\textbf{ELBO} (higher is better) and per pixel \textbf{reconstruction error} (lower is better). Our method achieves highest ELBO and lowest reconstruction error across all datasets.}
    \vspace{1mm}
    \begin{tabular}{|C|E|B|B|B|B|B|}
        \hline
        Metric                 & Dataset  & Normal                  & VampPrior            & GMM                  & Hierarchical         & Ours                 \\ \hline
        \multirow{3}{*}{ELBO}  & MNIST    & 1528.6 $\pm$ 21.3       & 1476.3 $\pm$ 19.4    & 1558.2 $\pm$ 24.4    & 1534.9 $\pm$ 19.8    & \textbf{1562.5} $\pm$ 25.1\\ \cline{2-7}
        & fashion                  & 1078.4 $\pm$ 23.4       & 1069.6 $\pm$ 22.1    & 1133.3 $\pm$ 23.9    & 1152.9 $\pm$ 24.9     &  \textbf{1174.0} $\pm$ 29.7   \\  \cline{2-7}
        & CelebA                          & 55564.8 $\pm$ 1751.3    & 56463.4 $\pm$ 1683.9 & 54162.5 $\pm$ 1716.5 & 56176.2 $\pm$ 1763.9  & \textbf{58596.1} $\pm$ 1815.3 \\ \hline\hline
        Per pixel       & MNIST      & 2.3 $\pm$ 0.06    & 2.5 $\pm$ 0.06    & 2.3 $\pm$ 0.05      & 2.4 $\pm$ 0.06     & \textbf{2.2} $\pm$ 0.06     \\  \cline{2-7}
        recons. error   & fashion    & 4.3 $\pm$ 0.12    & 4.1 $\pm$ 0.11    & 4.0 $\pm$ 0.11      & 3.7 $\pm$ 0.11     & \textbf{3.6} $\pm$ 0.11 \\  \cline{2-7}
        ($\times 0.01$) & CelebA     & 5.8 $\pm$ 0.21    & 5.9 $\pm$ 0.25    & 5.9 $\pm$ 0.20      & 5.7 $\pm$ 0.17     & \textbf{5.5} $\pm$ 0.18 \\ \hline
    \end{tabular}
    \label{tab:ELBO-recons-error}
\end{table}

\begin{figure}[t]
\centering
\includegraphics[width=0.9\textwidth]{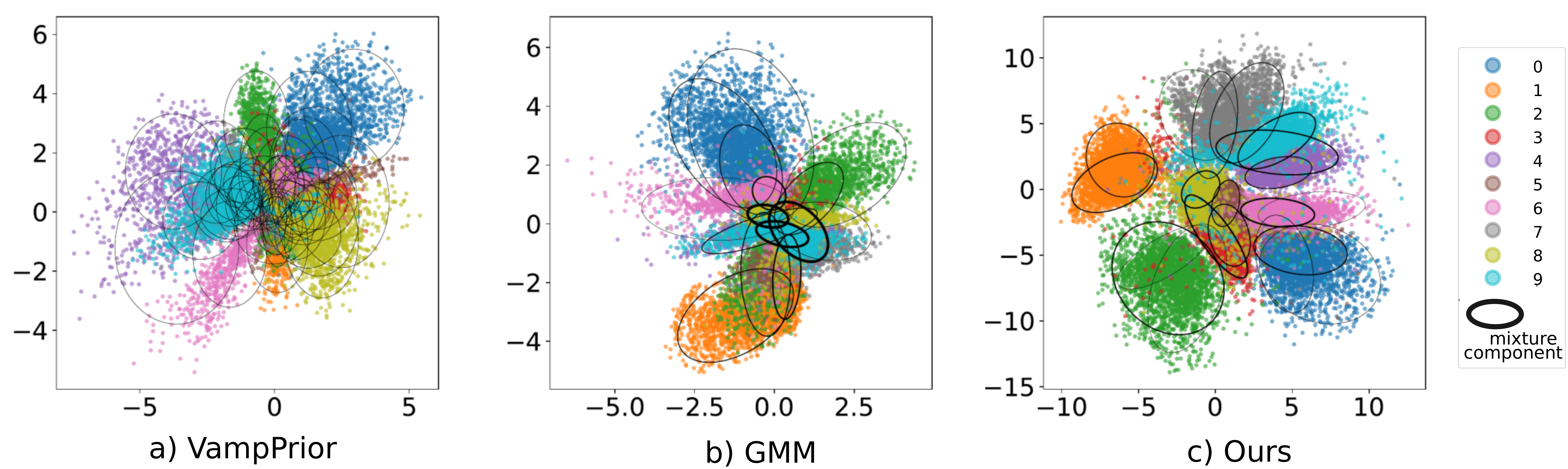}
    \vspace{-2mm}
    \caption{\textbf{The mixtures in our hyper prior naturally aligns with MNIST class labels}. We visualise encodings coloured by the digit labels for (a) VampPrior, (b) GMM prior and (c) our generative prior. Our method produces better clustering and the mixture components inferred in our method are naturally aligned with different classes.}
    \label{fig:mnist-clustering}
    \vspace{-1mm}
\end{figure}

Furthermore, our LaDDer model produces an embedding scheme that better preserves the semantics of the data. To see this, we visualise 30k latent encodings of the MNIST dataset in Fig.~\ref{fig:mnist-clustering}, where a VAE of 2D latent space is trained with a GMM prior, VampPrior and our generative prior respectively. All encodings are coloured by their class labels. Our method clearly gives better clustering results, where encodings of different classes are better separated. Furthermore, the mixture components estimated in our method are better aligned with class labels, whereas the mixtures from VampPrior and GMM prior have significant overlaps and do not have a consistent correspondence to specific class labels.  

\vspace{-2mm}
\subsection{Our Shortest Likely Path Traversal Gives Better Interpolation}
In Section \ref{sec:SLP}, we formulate the latent space interpolation between a pair of images as a shortest likely path (SLP) optimisation task, which favours the paths that go through high density regions of the inferred latent data distribution $p(\rvz)$ or $p(\rvt)$. Here we demonstrate our SLP method informed by the learnt latent distribution outperforms the conventional linear shortest path (SP) interpolation. 

\begin{figure}[t]
\centering
\includegraphics[width=0.95\textwidth]{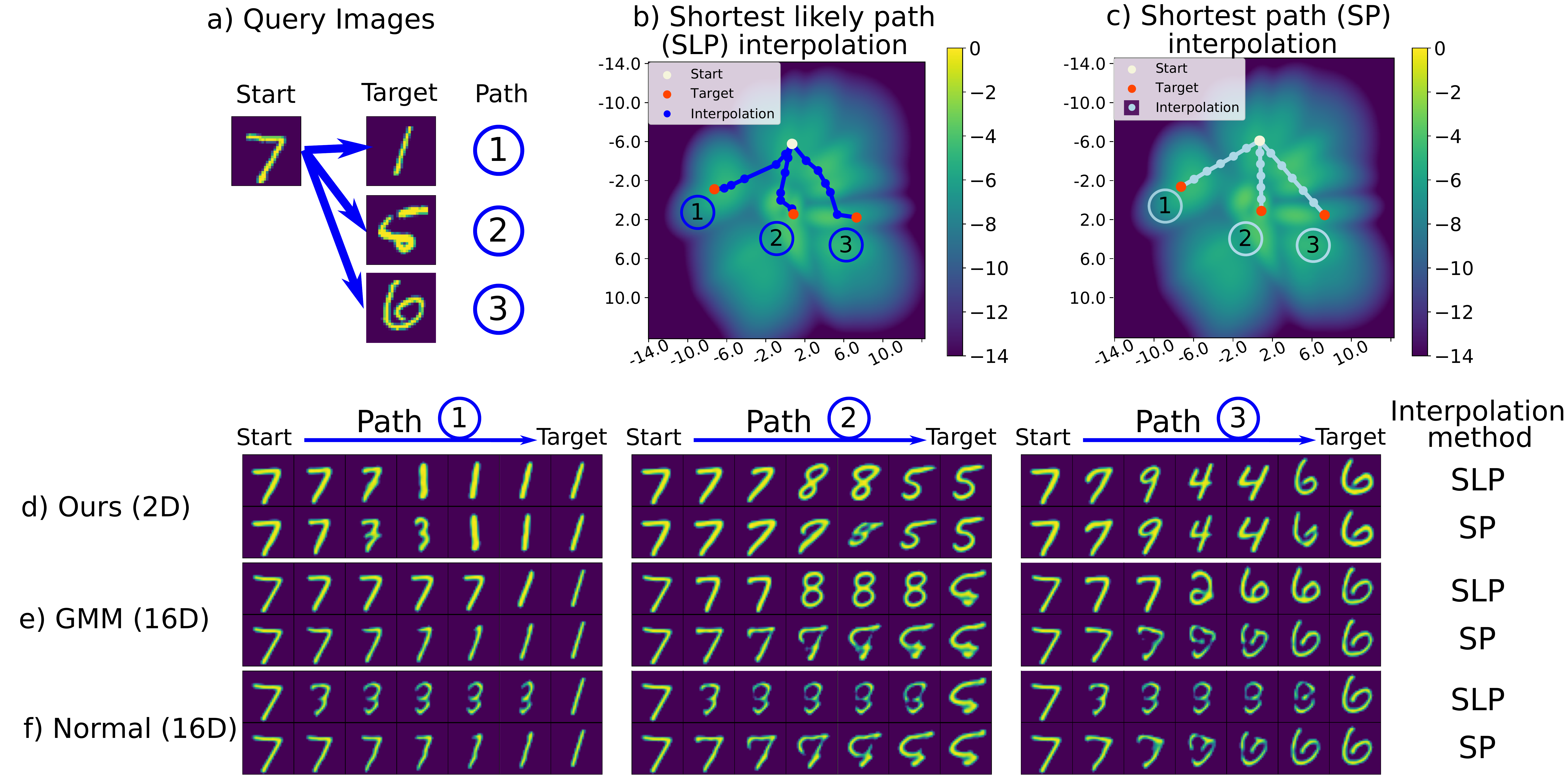}
    \vspace{-2mm}
    \caption{\textbf{Latent space interpolation for MNIST}. (a) The 3 pairs of MNIST images for the interpolation task. (b-c) The traversed paths produced by shortest likely path (SLP) and shortest path (SP) are visualised in our very low-dimensional $t$-space.
    (d-f) The interpolated images for both interpolation methods using the data distribution estimated by our method, a GMM prior and a normal prior.}
    \label{fig:mnist-traversal}
    \vspace{-3mm}
\end{figure}

Fig.~\ref{fig:mnist-traversal} illustrates the interpolation between 3 pairs of MNIST images through the data manifolds estimated by our method, a GMM prior and a normal prior. 
Notice that our LaDDer model can produce the same quality samples from a much lower dimension latent space compared to the VAE models with the GMM and the normal prior (2D vs 16D).
The super low latent dimension allows us to visualise the learnt data distribution $p(\rvt)$ and the traversed path in Fig.~\ref{fig:mnist-traversal}b-c. Our SLP interpolation results in paths that only step on regions with high likelihood of $p(\rvt)$, whereas SP interpolation ignores the topology of $p(\rvt)$ and often lands on the low density regions which do not correspond to realistic data samples, hence producing unrealistic samples (bottom rows in Fig.~\ref{fig:mnist-traversal}d-f).  
Notice that neither SLP nor SP gives a good interpolation for the normal prior model. This is because the normal prior poorly represents the latent data distribution and even if an encoding has a high likelihood wrt the normal prior, it does not correspond to a realistic data sample.
This reinforces the importance of obtaining an accurate modelling of the inferred data distribution in being able to utilise it for later tasks. 

\begin{figure}[t]
\centering
\includegraphics[width=0.9\textwidth]{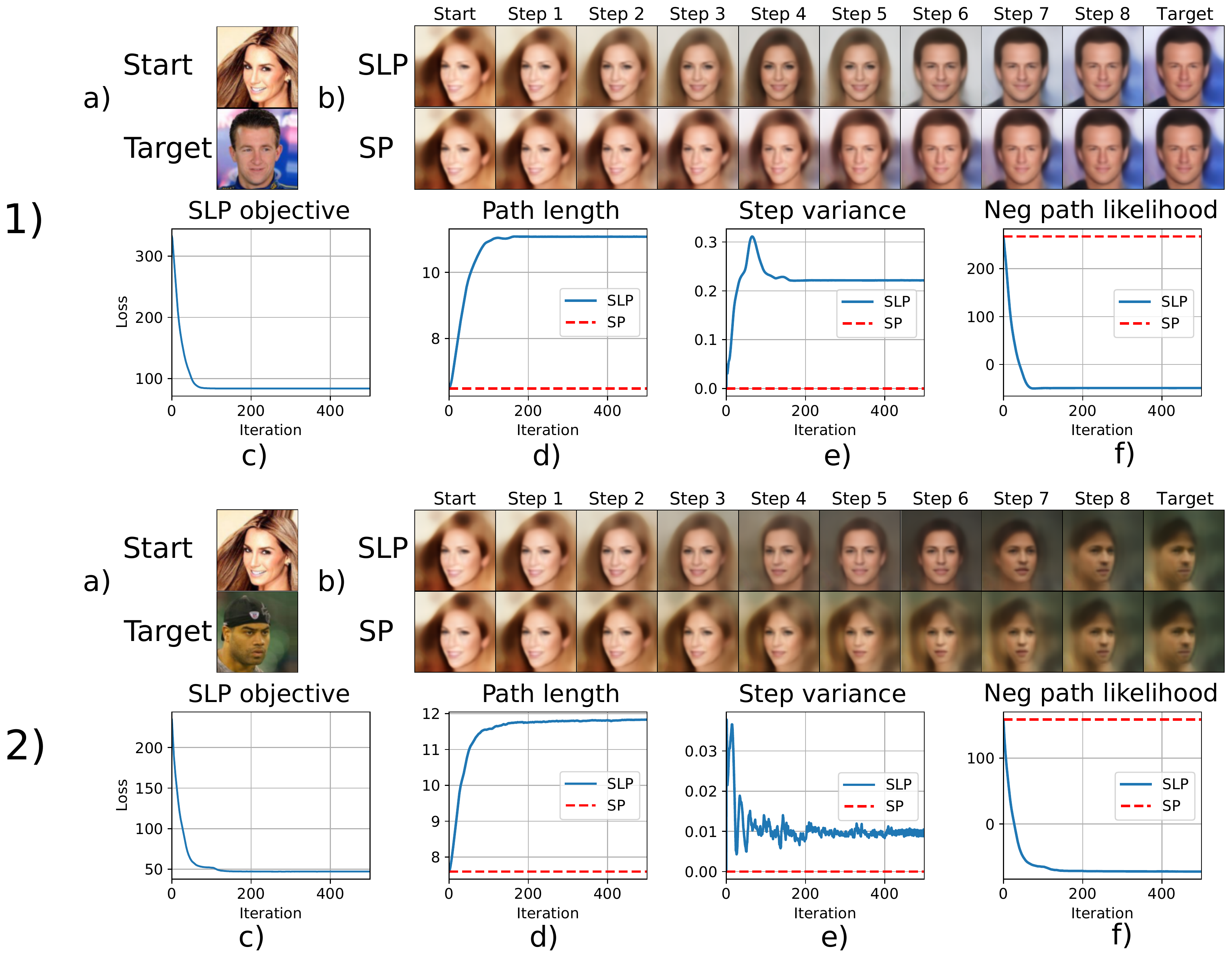}
    \vspace{-2mm}
    \caption{\textbf{Latent space interpolation for CelebA}. (a) The pair of images to be interpolated. (b) The sequence of interpolated images along the data manifold estimated by our LaDDer model, using two different interpolation methods. (c-f) Optimisation of different objectives in the our SLP objective. More examples are given in Supplementary Materials F4.}
    \label{fig:celeba-traversal}
    \vspace{-5mm}
\end{figure}

Fig.~\ref{fig:celeba-traversal} illustrates the interpolation between 2 pairs of CelebA images over the data manifold produced by our method, where the SLP optimisation takes place in the 32D latent space of the prior VAE. 
Here, we plot the different objectives in our SLP optimisation (blue line) for each example in (c-f). The overall objective ($\mathcal{O}_{\textrm{SLP}}$ in Eq (\ref{eq:SLP-obj})) smoothly converges in all examples. It is clear that our SLP solution trade-offs path length and allows different step lengths for obtaining high likelihood over the traversed path. As a result, our SLP method only produces realistic images which smoothly transform into the target face, whereas the SP method produces unrealistic faces along the traversed path.  

\vspace{-2mm}
\section{Conclusion and Future Work}
\label{sec:conclusion}
In this paper, we recognise the importance of adopting a sufficiently flexible prior in a VAE model to facilitate accurate modelling of the inferred latent data distribution. We propose LaDDer, which consists of multiple VAE models each acting on its predecessor’s latent representation and a non-parametric mixture as the hyper prior for the innermost VAE. From extensive experiments, we show that our method is able to accurately model latent data distribution of complex data. We also demonstrate how to use the derived latent distribution to facilitate further tasks, such as producing better interpolation along the derived data manifold. We believe that LaDDer can be helpful in estimating the data distribution for many challenging datasets and the derived data distribution can be useful for a wide range of applications. We will continue to explore along these directions. 

\section*{Acknowledgements}
We thank Stephen Roberts and Niki Trigoni for engaging discussions and supporting the research behind this project. Shuyu Lin is supported by the EPSRC Centre for Doctoral Training, EP/L015897/1, and the China Scholarship Council. Ronald Clark is supported by an Imperial College Research Fellowship. We thank the reviewers for helpful comments.

\appendix


\section*{Supplementary material}

# Supplementary Materials

In this document, we give additional information of the following topics related to our paper:

- the mathematical derivation for the re-arrangement of the cross-entropy term in the VAE ELBO loss (Section A),
- an algorithmic demonstration for our block coordinate ascent algorithm (Section B),
- the derivation for the reconstruction likelihood in VAE ELBO loss when a Laplace likelihood is used (Section C),
- and details of the datasets and pre-processing, experiment set-up, model architecture, ablation study and more results (Section D, E and F).

## A. Mathematical Derivations

Here we give details of the mathematical derivation for the re-arrangement of the cross-entropy term in the VAE ELBO loss, which is mentioned in Section 2 and 3.1 of the main texts.

### A.1. Re-arrangement of cross-entropy wrt prior

The cross-entropy term $\mathbb{E}_{p_\mathcal{D}(\mathbf{x})}\mathbb{E}_{q_\phi(\mathbf{z}|\mathbf{x})}[\log p(\mathbf{z})]$ appears as ③ in Eq (2) in the main texts. It is commonly used as a regulariser to the sample posteriors $q_\phi(\mathbf{z}|\mathbf{x})$. With the following re-arrangement, we show that this term is in fact an alignment requirement between the prior distribution $p(\mathbf{z})$ and the inferred aggregate posterior $q_\phi(\mathbf{z})$, i.e. the population distribution of all the samples in the learnt latent space. The re-arrangement is done as follows:

$$\mathbb{E}_{p_\mathcal{D}(\mathbf{x})}\mathbb{E}_{q_\phi(\mathbf{z}|\mathbf{x})}[\log p(\mathbf{z})] = \int_\mathbf{z} \int_\mathbf{x} \log p(\mathbf{z}) q_\phi(\mathbf{z}|\mathbf{x}) p_\mathcal{D}(\mathbf{x}) \, d\mathbf{x} \, d\mathbf{z} \quad (1)$$

$$= \int_\mathbf{z} \log p(\mathbf{z}) \left( \int_\mathbf{x} q_\phi(\mathbf{z}|\mathbf{x}) p_\mathcal{D}(\mathbf{x}) \, d\mathbf{x} \right) d\mathbf{z} \quad (2)$$

$$= \int_\mathbf{z} \log p(\mathbf{z}) \, \mathbb{E}_{p_\mathcal{D}(\mathbf{x})}[q_\phi(\mathbf{z}|\mathbf{x})] \, d\mathbf{z} \quad (3)$$

$$= \int_\mathbf{z} \log p(\mathbf{z}) \, q_\phi(\mathbf{z}) d\mathbf{z} \quad (4)$$

$$= \mathbb{E}_{q_\phi(\mathbf{z})}[\log p(\mathbf{z})], \quad (5)$$

where

$$\text{aggregate posterior } q_\phi(\mathbf{z}) = \int q_\phi(\mathbf{z}|\mathbf{x}) p_\mathcal{D}(\mathbf{x}) d\mathbf{x} = \mathbb{E}_{p_\mathcal{D}(\mathbf{x})}[q_\phi(\mathbf{z}|\mathbf{x})] \approx \frac{1}{N} \sum_{i=1}^{N} q_\phi(\mathbf{z}|\mathbf{x}_i). \quad (6)$$

When a cross-entropy is maximised between two distributions, the two distributions will try to have more overlaps and the optimal solution will be reached when the two distributions become the same. Therefore, this cross-entropy term in the ELBO objective simply implies that the optimal prior should be matching the aggregate posterior.

### A.2. Re-arrangement of the Cross-entropy between Hyper Prior and Aggregate Hyper Posterior

A similar re-arrangement for the cross-entropy term in the prior VAE ELBO (Eq (4) in the main texts) can be also achieved in the following equations. This result indicates that the optimal hyper prior $p(\mathbf{t})$ should be chosen to best fit the aggregate hyper posterior $q_{\beta,\phi}(\mathbf{t})$, which is defined as an average encoding of all possible data encodings projected to the $t$-space. This fitting is crucial and failure to do so will result in poor modelling performance as shown in Section 4.1 of the main texts

for the hierarchical model.

$$\mathbb{E}_{p_\mathcal{D}(\mathbf{x})}\mathbb{E}_{q_\phi(\mathbf{z}|\mathbf{x})}\mathbb{E}_{q_\beta(\mathbf{t}|\mathbf{z})}[\log p(\mathbf{t})] = \mathbb{E}_{q_{\beta,\phi}(\mathbf{t})}[\log p(\mathbf{t})] \quad (7)$$

$$= \int_\mathbf{t}\int_\mathbf{z}\int_\mathbf{x} \log p(\mathbf{t}) q_\beta(\mathbf{t}|\mathbf{z}) q_\phi(\mathbf{z}|\mathbf{x}) p_\mathcal{D}(\mathbf{x})\,d\mathbf{x}\,d\mathbf{z}\,d\mathbf{t} \quad (8)$$

$$= \int_\mathbf{t}\int_\mathbf{z} \log p(\mathbf{t}) q_\beta(\mathbf{t}|\mathbf{z}) \left(\int_\mathbf{x} q_\phi(\mathbf{z}|\mathbf{x}) p_\mathcal{D}(\mathbf{x})\,d\mathbf{x}\right) d\mathbf{z}\,d\mathbf{t} \quad (9)$$

$$= \int_\mathbf{t} \log p(\mathbf{t}) \left(\int_\mathbf{z} q_\beta(\mathbf{t}|\mathbf{z}) q_\phi(\mathbf{z}) d\mathbf{z}\right) d\mathbf{t} \quad (10)$$

$$= \int_\mathbf{t} \log p(\mathbf{t}) q_{\phi,\beta}(\mathbf{t})\,d\mathbf{t} \quad (11)$$

$$= \mathbb{E}_{q_{\phi,\beta}(\mathbf{t})}[\log p(\mathbf{t})], \quad (12)$$

where

$$\text{aggregate hyper posterior } q_{\phi,\beta}(\mathbf{t}) = \int q_\beta(\mathbf{t}|\mathbf{z}) q_\phi(\mathbf{z}) d\mathbf{z} = \mathbb{E}_{q_\phi(\mathbf{z})}[q_\beta(\mathbf{t}|\mathbf{z})] = \mathbb{E}_{q_\phi(\mathbf{z})}[q_\beta(\mathbf{t}|\mathbf{z})]. \quad (13)$$

## B. Block Coordinate Ascent Algorithm

Here we give an algorithmic demonstration for our block coordinate ascent algorithm, which is introduced in Section 3.3 of the main texts as our optimisation strategy for all the model parameters in our LaDDer model (i.e. the data VAE parameters $\theta, \phi$, the prior VAE parameters $\alpha, \beta$ and the GMM hyper prior $\xi$).

---
**Algorithm 1** Block coordinate ascent to optimise model parameters $\theta, \phi, \alpha, \beta, \xi$
---

$\phi, \theta\ \alpha\ \beta \leftarrow$ Initialize parameters
$\xi \leftarrow$ Initialize as unit Gaussian
**repeat** until convergence of $\mathcal{L}''(\mathbf{x}; \theta, \phi, \alpha, \beta, \xi)$
$\quad \theta, \phi \leftarrow$ Update by optimising $\mathcal{L}''(\mathbf{x}; \theta, \phi\,|\,\alpha, \beta, \xi)$ using AdamOptimiser
$\quad \alpha, \beta \leftarrow$ Update by optimising $\mathcal{L}'(\mathbf{z}; \alpha, \beta\,|\,\xi, \phi)$ using AdamOptimiser
$\quad \xi \leftarrow$ At the end of an epoch, update by optimising $\mathcal{L}(\mathbf{t}; \xi)$ using the updates in coordinate ascent (Blei et al., 2006)
**return** $\theta, \phi, \alpha, \beta, \xi$

---

## C. Laplace Distribution to Model the Decoder Likelihood

Throughout our experiments, we use a Laplace distribution to model the reconstruction likelihood $p_\theta(\boldsymbol{x}_i|\mathbf{z})$ (i.e. ① in Eq (2) from the main texts). Here we give the mathematical details of this distribution and it will become clear that this choice of the decoder distribution results in a L1 reconstruction error loss in the overall learning objective, which is more suited for modelling image data. The derivation follows the idea proposed in (Lin et al., 2019).

The Laplace likelihood is defined as

$$p_\theta(\boldsymbol{x}_i|\mathbf{z}) = \prod_{k=1}^{d} \frac{1}{2\sigma} \exp -\frac{|\boldsymbol{x}_i^k - \mu_\theta^k(\mathbf{z})|}{\sigma}, \quad (14)$$

where $\sigma$ is a hyper parameter which we also optimise under the ELBO objective. Substituting the above equation into the negative ELBO loss in Eq (2) of the main texts, we have

$$-\mathcal{L}(\mathbf{x}; \theta, \phi, \sigma) = \frac{1}{\sigma}\mathbb{E}_\mathbf{z}\underbrace{\left[\sum_{k=1}^{d} |\mu_\theta^k(\mathbf{z}) - \boldsymbol{x}_i^k|\right]}_{①} + D_{\mathrm{KL}}\big[q_\phi(\mathbf{z}|\boldsymbol{x}_i)\,\|\,p(\mathbf{z})\big] + d \log 2\sigma, \quad (15)$$

where $\mathbb{E}_{p_\mathcal{D}(\mathbf{x})}[\cdot]$ is omitted. Now it is clear that the reconstruction likelihood ① simplifies to an element-wise absolute error (or L1 norm) also labelled as ① here.

## D. Datasets and Pre-processing

We ran experiments on MNIST, Fashion MNIST and CelebA datasets. For all datasets, we take the images as real-valued data. Here we give details of these datasets and how we pre-process them for our training the VAE models in Section 4 of the main texts.

### D.1. MNIST and Fashion MNIST

MNIST and Fashion MNIST datasets both consist of a training and a test set with 60k and 10k images of hand-written digits/clothes, where each image is a 28×28 grey-scale image. We rescale the original images so that the pixel values are within the range $[0, 1]$.

### D.2. CelebA

CelebA dataset consists of 202599 RGB images of celebrity faces. We first resize all the images to $128 \times 128$ and rescale the pixel values to be within $[0, 1]$. We then randomly select 20k images (about 10%) to be the test set.

## E. Model Architectures and Training Parameters

### E.1. MNIST and Fashion MNIST

We use CNN architectures for both MNIST and Fashion MNIST encoder and decoder network (with additional depth to space operations for the decoder), as detailed below. We use the exactly same network for both datasets.

Encoder network:

$$\boldsymbol{x} \in \mathcal{R}^{28 \times 28 \times 1} \to \text{Symmetric padding} \to \mathcal{R}^{32 \times 32 \times 1} \to \text{Conv}_{16}^{3\times 3} \ (\text{stride}=2) \to \text{ReLU}$$
$$\to \text{Conv}_{64}^{3\times 3} \ (\text{stride}=2) \to \text{ReLU}$$
$$\to \text{Conv}_{256}^{3\times 3} \ (\text{stride}=2) \to \text{ReLU}$$
$$\to \text{FC}_{64} \to \text{FC}_{d_z} \Rightarrow \mu \in \mathcal{R}^{d_z}$$
$$\searrow$$
$$\text{FC}_{d_z} \to \text{ReLU} \Rightarrow \sigma \in \mathcal{R}^{d_z}$$

Decoder network:

$$\boldsymbol{z} = \mu + \epsilon \odot \sigma \to \text{FC}_{4096} \to \text{ReLU} \to \text{Reshape} \to \mathcal{R}^{1\times 1\times 4096}$$
$$\to \text{depth to space}_4 \to \mathcal{R}^{4\times 4\times 256} \to \text{Conv}_{256}^{3\times 3} \ (\text{stride}=1) \to \text{ReLU}$$
$$\to \text{depth to space}_2 \to \mathcal{R}^{8\times 8\times 64} \to \text{Conv}_{64}^{3\times 3} \ (\text{stride}=1) \to \text{ReLU}$$
$$\to \text{depth to space}_2 \to \mathcal{R}^{16\times 16\times 16} \to \text{Conv}_{16}^{3\times 3} \ (\text{stride}=1) \to \text{ReLU}$$
$$\to \text{depth to space}_2 \to \mathcal{R}^{32\times 32\times 4} \to \text{Conv}_{1}^{5\times 5} \ (\text{stride}=1, \text{padding=valid})$$
$$\to \text{Sigmoid} \to \text{Reshape} \Rightarrow \hat{\boldsymbol{x}} \in \mathcal{R}^{28\times 28\times 1}$$

Prior VAE network:

$$\boldsymbol{z} \to (\text{FC}_{512} \to \text{Leaky ReLU}) \times 3 \to \text{FC}_{d_t} \to \mu_t \in \mathcal{R}^{d_t}$$
$$\searrow$$
$$\text{FC}_{d_t} \to \text{ReLU} \Rightarrow \sigma_t \in \mathcal{R}^{d_t}$$
$$\boldsymbol{t} = \mu_t + \epsilon \odot \sigma_t \to (\text{FC}_{512} \to \text{Leaky ReLU}) \times 3$$
$$\to \text{FC}_{d_z} \to \hat{z} \in \mathcal{R}^{d_z}$$

$d_z$ is the dimension of $z$-space, $d_t$ is the latent dimension of our prior VAE, $\mu$ is code mean and $\sigma$ is decoder variance parameter introduced in Section C, $\epsilon \sim \mathcal{N}(0, \mathcal{I})$ and unless otherwise specified the padding in the convolution operation is 'same'. We use the following hyper-parameters to train the network:

| Batch size | $d_z$ | $d_t$ | N epoch | Optimizer | Learning rate | Padding |
|---|---|---|---|---|---|---|
| 256 | 16 | 2 | 50 | Adam | $5 \times 10^{-4}$ | SAME |

### E.2. CelebA

For CelebA dataset, we implement a CNN encoder and a Style-GAN generator, as detailed below. Encoder network:

$$x \in \mathcal{R}^{128 \times 128 \times 3} \to \text{Conv}^{3\times3}_{32} \ (\text{stride}=2) \to \text{Leaky ReLU}$$
$$\to \text{Conv}^{3\times3}_{64} \ (\text{stride}=2) \to \text{Leaky ReLU}$$
$$\to \text{Conv}^{3\times3}_{128} \ (\text{stride}=2) \to \text{Leaky ReLU}$$
$$\to \text{Conv}^{3\times3}_{256} \ (\text{stride}=2) \to \text{Leaky ReLU}$$
$$\to \text{Conv}^{3\times3}_{512} \ (\text{stride}=2) \to \text{Leaky ReLU}$$
$$\to \text{Conv}^{3\times3}_{1024} \ (\text{stride}=2, \text{padding=valid}) \to \text{Leaky ReLU}$$
$$\to \text{FC}_{512} \to \text{FC}_{d_z} \Rightarrow \mu \in \mathcal{R}^{d_z}$$
$$\searrow$$
$$\text{FC}_{d_z} \Rightarrow \sigma \in \mathcal{R}^{d_z}$$

Decoder network which is inspired by the Style-GAN architecture is given below. The Style Module refers to a fully connected layer that first doubles the channel dimensions and then merges the channel dimension back to the original dimension by summing the first half channels with the original channel value weighted second half channels. This operation is modified from the original Style-GAN generator to fit with a VAE model.

$$z = \mu + \epsilon \odot \sigma \to (\text{FC}_{512} \to \text{Leaky ReLU}) \times 8 \to \text{Reshape} \to \mathcal{R}^{1 \times 1 \times 512}$$
$$\to \text{Resize}^{2\times2}_{128} \to \text{CNN}^{3\times3}_{512} \to \text{Instance Norm} \to \text{Style Module} \to \mathcal{R}^{2\times2\times512}$$
$$\to \text{Resize}^{4\times4}_{128} \to \text{CNN}^{3\times3}_{512} \to \text{Instance Norm} \to \text{Style Module} \to \mathcal{R}^{4\times4\times512}$$
$$\to \text{Resize}^{8\times8}_{128} \to \text{CNN}^{3\times3}_{512} \to \text{Instance Norm} \to \text{Style Module} \to \mathcal{R}^{8\times8\times512}$$
$$\to \text{Resize}^{16\times16}_{128} \to \text{CNN}^{3\times3}_{256} \to \text{Instance Norm} \to \text{Style Module} \to \mathcal{R}^{16\times16\times256}$$
$$\to \text{Resize}^{32\times32}_{64} \to \text{CNN}^{3\times3}_{256} \to \text{Instance Norm} \to \text{Style Module} \to \mathcal{R}^{32\times32\times256}$$
$$\to \text{Resize}^{64\times64}_{64} \to \text{CNN}^{3\times3}_{128} \to \text{Instance Norm} \to \text{Style Module} \to \mathcal{R}^{64\times64\times128}$$
$$\to \text{Resize}^{128\times128}_{32} \to \text{CNN}^{3\times3}_{128} \to \text{Instance Norm} \to \text{Style Module} \to \mathcal{R}^{128\times128\times128}$$
$$\to \text{CNN}^{3\times3}_{128} \to \text{Leaky ReLU} \to \text{CNN}^{3\times3}_{3} \to \mathcal{R}^{128\times128\times3}$$

Prior VAE network:

$$z \to (\text{FC}_{512} \to \text{Leaky ReLU}) \times 5 \to \text{FC}_{d_t} \to \mu_t \in \mathcal{R}^{d_t}$$
$$\searrow$$
$$\text{FC}_{d_t} \to \text{ReLU} \Rightarrow \sigma_t \in \mathcal{R}^{d_t}$$
$$t = \mu_t + \epsilon \odot \sigma_t \to (\text{FC}_{512} \to \text{Leaky ReLU}) \times 5$$
$$\to \text{FC}_{d_z} \to \hat{z} \in \mathcal{R}^{d_z}$$

We use the following hyper-parameters to train the network:

| Batch size | $d_z$ | $d_t$ | N epoch | Optimizer | Learning rate | Padding |
|---|---|---|---|---|---|---|
| 256 | 256 | 32 | 75 | Adam | $5 \times 10^{-4}$ | SAME |

## F. More Experimental Results

Now we show more results that supplement the results from the main texts. We give more generated samples from each VAE model, FID scores on the MNIST and fashion-MNIST datasets, an ablation study on the impacts of the prior VAE's latent dimension on the overall modelling performance and finally more examples of latent space interpolation using the derived latent distribution.

**Supplementary Materials for LaDDer: Latent Data Distribution Modelling with a Generative Prior**

**F.1. Generated Samples from Inferred Prior Distribution**

Here we generate samples from the inferred prior distribution of each VAE model. The sample quality indicates how well the prior distribution matches the learnt latent data distribution. As can be seen, our method and GMM prior produce the best quality samples. However, in the higher dimensional latent space which is often needed for more complex datasets such as CelebA, our method outperforms the GMM prior. It is also very clear that the standard VAE with a normal prior fails to represent the inferred data distribution at all. This calls for caution in using the unit Gaussian prior if the inferred latent distribution is actually needed to represent the realistic data samples in later tasks.

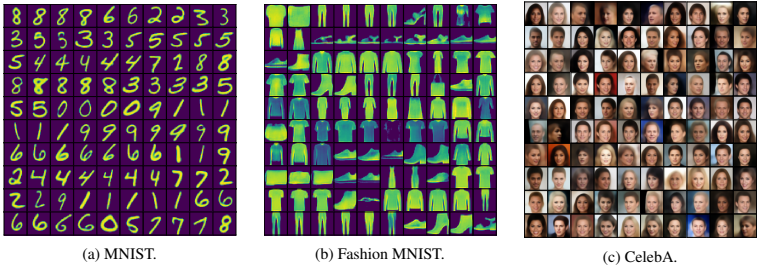

(a) MNIST.     (b) Fashion MNIST.     (c) CelebA.

*Figure 1.* 100 images generated from the inferred prior distribution in our LaDDer model for each of the 3 datasets.

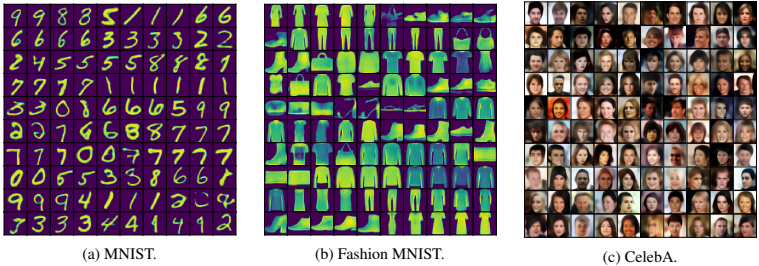

(a) MNIST.     (b) Fashion MNIST.     (c) CelebA.

*Figure 2.* 100 images generated from the inferred prior distribution in a GMM-VAE model for each of the 3 datasets.

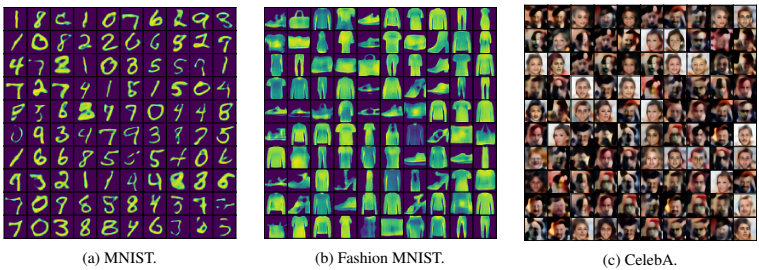

(a) MNIST.     (b) Fashion MNIST.     (c) CelebA.

*Figure 3.* 100 images generated from the inferred prior distribution in VampPrior model for each of the 3 datasets.

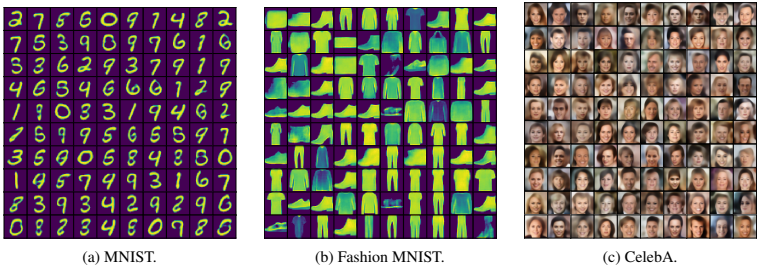

Figure 4. 100 images generated from the inferred prior distribution in a VAE with a hierarchical prior for each of the 3 datasets.

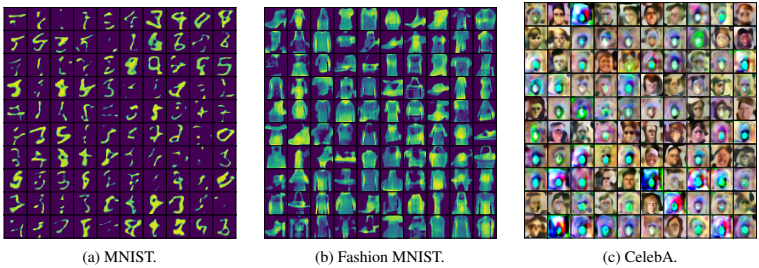

Figure 5. 100 images generated from the inferred prior distribution in standard VAE (normal prior) for each of the 3 datasets.

### F.2. FID Scores on MNIST and fashion-MNIST

In Section 4.1 from the main texts, we demonstrate that our generative prior can accurately model the inferred latent data distribution by evaluating the FID scores of both generated and reconstructed samples. We pay special attention to the difference in the visual quality between the generation and reconstruction. This is because the reconstruction samples quality of a VAE indicates the best sample quality the VAE can generate (if the prior 100% matches the inferred data manifold point-by-point). We give the FID score for CelebA dataset in the main texts. Here we also give the FID scores for MNIST and fashion-MNIST datasets, which tell a very similar story that our LaDDer model can generate very good quality samples and the gap between generation and reconstruction is small.

Table 1. FID scores (lower is better) of generated and reconstructed samples from 5 VAE models with different priors. Our generative prior and GMM hyper prior achieves best sample quality in both generation and reconstruction and the minimal gap between the two.

| Dataset | Prior model | Normal | VampPrior | GMM | Hierarchical | Ours | Real image |
|---|---|---|---|---|---|---|---|
| MNIST digit | Generation | 401.4 ± 4.4 | 47.2 ± 3.6 | 2.3 ± 0.5 | 52.4 ± 4.7 | 6.4 ± 1.3 | 2.3 ± 0.4 |
|  | Reconstruction | 2.1 ± 0.8 | 1.4 ± 0.4 | 2.5 ± 0.8 | 5.4 ± 1.4 | 2.4 ± 0.4 |  |
|  | Difference | 399.3 ± 5.2 | 45.8 ± 4.0 | -0.2 ± 1.3 | 47.0 ± 3.3 | 4.0 ± 1.7 |  |
| MNIST fashion | Generation | 266.1 ± 2.4 | 23.6 ± 1.2 | 6.0 ± 1.0 | 243.8 ± 2.9 | 5.2 ± 0.9 | 0.4 ± 0.1 |
|  | Reconstruction | 3.2 ± 0.4 | 3.0 ± 0.2 | 2.9 ± 0.4 | 1.8 ± 0.2 | 1.0 ± 0.4 |  |
|  | Difference | 262.9 ± 2.8 | 20.6 ± 1.4 | 3.1 ± 1.4 | 242.0 ± 3.1 | 4.2 ± 1.3 |  |

An interesting effect shown here is that the VAE model with a GMM prior also achieves very good performance. We reckon this is because the MNIST datasets are relatively simple and hence a GMM fitting in a relatively low-dimensional $z$-space can match the overall data distribution fairly well. However, when the dataset becomes more complex, such as the CelebA dataset, our generative prior is more flexible and outperforms the GMM prior, as shown in the main texts Section 4.1.

### F.3. Ablation Study: the Impact of Hyper Latent Dimension on Performance

Here we carry out an ablation study on the impact of the latent dimension of our prior VAE model on the overall modelling performance, specifically on the generation quality. Over all three datasets, the generation quality improves as we allow the

**Supplementary Materials for LaDDer: Latent Data Distribution Modelling with a Generative Prior**

prior VAE to adopt a higher latent dimension. However, the benefit of increasing the latent dimension diminishes as the dimension reaches certain level. This indicates that the dimension reduction achieved by the prior VAE module is still valid and does not limit the model's performance as long as its latent dimension is not set to be overly small.

Table 2. FID scores (lower is better) of generated samples from our generative prior with different latent dimension $d_t$. It is clear with higher latent dimension, the generation improves. However, the benefits diminish as the dimension increases.

| Dataset | Ours | | | | | Real image |
|---|---|---|---|---|---|---|
| | 2d | 4d | 8d | 16d | 32d | |
| MNIST digit | 58.7 ± 1.8 | 35.1 ± 8.1 | 6.4 ± 1.3 | 6.9 ± 0.8 | N/A | 2.3 ± 0.4 |
| MNIST fashion | 25.3 ± 1.6 | 9.9 ± 1.2 | 5.5 ± 0.7 | 5.2 ± 0.9 | N/A | 0.4 ± 0.1 |
| CelebA | 175.7 ± 0.5 | 148.4 ± 0.6 | 138.4 ± 0.5 | 133.5 ± 0.6 | 132.7 ± 0.5 | 5.9 ± 0.5 |

**F.4. Latent Space Interpolation Through Our Shortest Likelihood Path**

In the main texts, we demonstrate how to use the estimated latent distribution to facilitate further tasks through a latent space interpolation example. We show that the derived latent distribution can inform the interpolation to only step on high likelihood regions in the latent space. As a result, the inferred samples remain realistic throughout the path and render smooth transitions into the target image. Here, we show more examples on the interpolation task and we also show our shortest likely path traversal on two other VAE models to indicate it can be generally applied to other methods. Across the interpolation results, the interpolation informed with the learnt latent distribution generates better samples. The exception is the VAE with a normal prior and this is because the normal prior poorly models the true data distribution.

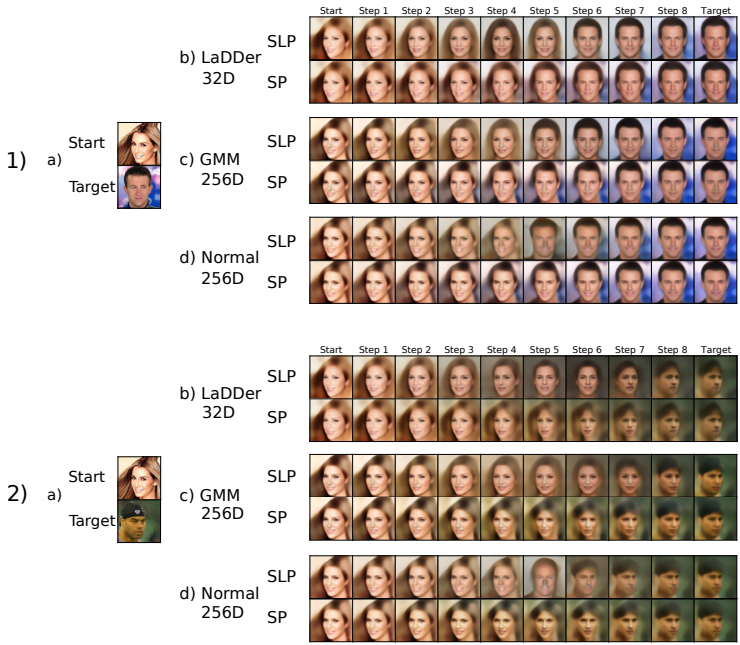

Figure 6. Latent space interpolation between a pair of CelebA images. We show the interpolated images for 3 VAE models: (b) our proposed LaDDer, (c) a GMM prior and (d) a normal prior. For each method, top row gives the interpolation of our shortest likely path (SLP) method and bottom row gives the interpolation of the shortest path (SP) method.

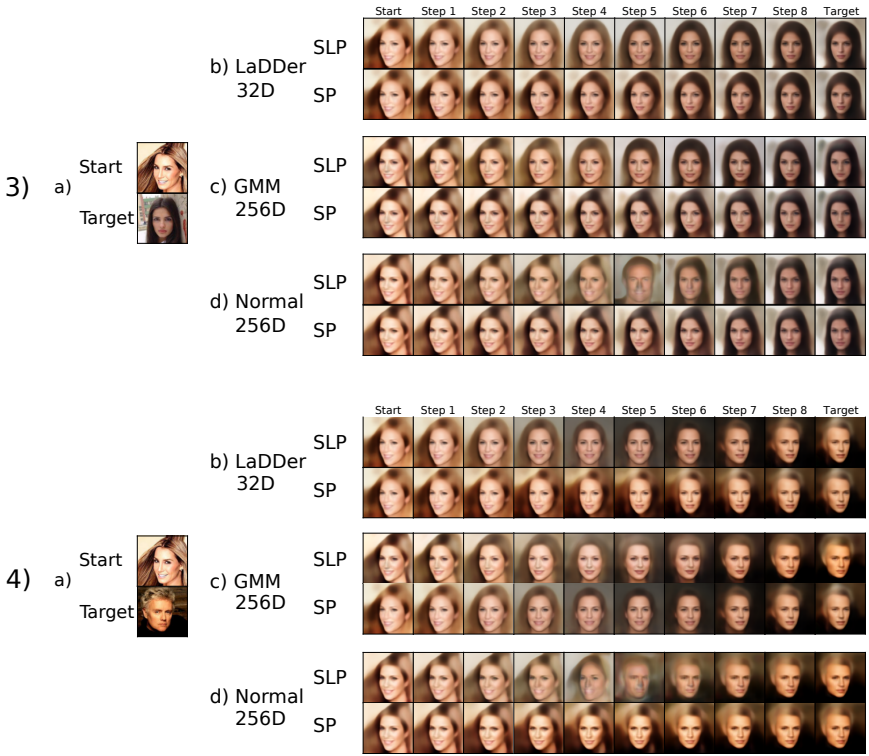

*Figure 7.* Latent space interpolation between a pair of CelebA images. We show the interpolated images for 3 VAE models: (b) our proposed LaDDer, (c) a GMM prior and (d) a normal prior. For each method, top row gives the interpolation of our shortest likely path (SLP) method and bottom row gives the interpolation of the shortest path (SP) method.



\end{document}